\documentclass[english]{article}
\usepackage[T1]{fontenc}
\usepackage[latin9]{inputenc}
\usepackage{array}
\usepackage{float}
\usepackage{amsmath}
\usepackage{amsthm}
\usepackage{amssymb}
\usepackage{graphicx}
\usepackage{wasysym}

\makeatletter

\providecommand{\tabularnewline}{\\}

\numberwithin{equation}{section}
\numberwithin{figure}{section}

\makeatother

\usepackage{babel}
\begin{document}
\title{Event Driven Clustering Algorithm}
\author{David El-Chai Ben-Ezra, Adar Tal, and Daniel Brisk}
\maketitle
\begin{abstract}
This paper introduces a novel asynchronous, event-driven algorithm
for real-time detection of small event clusters in event camera data.
Similar to hierarchical agglomerative clustering methods, the proposed
algorithm detects clusters based on their spatio-temporal proximity.
However, it explicitly leverages the asynchronous structure of event
camera data and employs a simple yet efficient decision mechanism,
achieving a linear time complexity of $\Theta(N)$, where $N$ is
the number of events. Furthermore, the runtime is independent of the
sensor resolution, i.e., the number of pixels.
\end{abstract}
Key words: event camera, neuromorphic vision, asynchronous algorithms,
hierarchical clustering, real-time detection.

\section{Introduction}

Event cameras represent a paradigm shift in visual sensing. Unlike
conventional cameras that sample the scene at a fixed frame rate,
each pixel in an event camera operates independently and asynchronously,
generating an \textquotedblleft event\textquotedblright{} whenever
it detects a change in brightness. This sensing mechanism enables
the capture of extremely fast motion while providing several intrinsic
advantages, including high dynamic range and microsecond-level temporal
resolution. Due to these properties, event cameras have the potential
to significantly improve a wide range of applications, including object
detection and classification, tracking, motion segmentation, optical
flow estimation, 3D reconstruction, pose estimation, and simultaneous
localization and mapping (SLAM) \cite{key-1,key-2}. 

Despite these advantages, effectively exploiting event camera data
remains challenging. The asynchronous and sparse nature of the data
requires the development of novel processing methods tailored to this
sensing paradigm (see \cite{key-3-1}-\cite{key-7}). In addition,
algorithms must efficiently handle high-resolution, temporally ordered
streams of events. Consequently, considerable effort has been devoted
to developing event-based methods (see, e.g. \cite{key-8-1}-\cite{key-20}).
In particular, several asynchronous clustering approaches for event-based
data have been proposed (see \cite{key-21}-\cite{key-25}).

In this paper, we propose a simple yet powerful asynchronous clustering
algorithm for detecting small objects. The algorithm operates in linear
time complexity $\Theta(N)$, depending only on the number of events
and not on the number of clusters or the spatial resolution of the
sensor (see $\varoint$\ref{sec:Algorithm-description}). This distinguishes
it from both classical hierarchical clustering methods and existing
asynchronous approaches, making it especially suitable for real-time
applications. Our method exploits the temporal ordering of event streams
and builds upon the concept of time surfaces (\cite{key-1}, \cite{key-26}-\cite{key-31}).
It processes events sequentially, making clustering decisions online
without revisiting past events. Notably, the algorithm avoids temporal
binning and instead performs all decisions during a single pass through
the event stream.

The algorithm is designed to detect small, localized signals that
naturally form clusters originating from a \textquotedblleft root\textquotedblright{}
event (see $\varoint$\ref{sec:Data-interpretation-and}) . A key
advantage is that the root of a cluster is identified immediately
once the cluster exceeds a predefined size threshold---well before
the clustering process is complete. Similar to other spatio-temporal
clustering methods, the algorithm relies on two main criteria: the
spatio-temporal proximity between events and a predefined threshold
on the cluster size.

\section{Data Representation and Algorithm Output\label{sec:Data-interpretation-and}}

\subsection{Data Structure of Event Cameras}

We begin with a brief overview of the operating principles of event
cameras. Each pixel maintains a reference illumination level and generates
an event whenever the change in brightness relative to this reference
exceeds a predefined threshold. Upon detecting such a change, the
pixel performs two actions:
\begin{itemize}
\item It updates its reference value to the current illumination level. 
\item It emits an event containing the following information:
\begin{itemize}
\item The timestamp of the event (with microsecond resolution), 
\item The spatial coordinates of the pixel, 
\item The polarity of the event, indicating whether the brightness change
is positive or negative.
\end{itemize}
\end{itemize}
Consequently, instead of producing image frames, the camera outputs
an asynchronous stream of events ordered by time. Each event is represented
as 

\[
v_{i}=(t_{i},x_{i},y_{i},p_{i})
\]
where $t_{i}$ denotes the timestamp, $(x_{i},y_{i})$ are the pixel
coordinates, and $p_{i}$ is the polarity, with $p_{i}=1$ for positive
events and $p_{i}=-1$ for negative events. In the sequel, the polarity
$p_{i}$ is not used and will therefore be omitted.

\subsection{Graph-Theoretic Interpretation}

The output of the algorithm can be naturally formulated by interpreting
the event camera output---serving as the input to the algorithm---within
a graph-theoretic framework. Let $\delta>0$ and $d\in\mathbb{N}$,
and consider the set of events $V=\{v_{i}\}_{i=1}^{N}$ as the vertices
of a directed graph $G=(V,E)$. The edge set $E$ is defined inductively
with respect to the event index $i$ as follows. 

Let $G_{k}=(V_{k},E_{k})$ denote the subgraph induced by the first
$k$ events, $\{v_{i}\}_{i=1}^{k}$, and consider the next event $v_{k+1}=(t_{k+1},x_{k+1},y_{k+1})$. 
\begin{enumerate}
\item If there exists a connected component $T=(V_{T},E_{T})$ of $G_{k}$,
such that:
\begin{itemize}
\item there exists $v=(t,x,y)\in V_{T}$ with $(x,y)=(x_{k+1},y_{k+1})$,
and
\item $t_{k+1}-t_{\max\left\{ j\,|\,v_{j}\in V_{T}\right\} }\leq\delta$, 
\end{itemize}
then we define
\[
G_{k+1}:=(V_{k}\cup\{v_{k+1}\},E_{k}\cup\{(v_{\min\left\{ j\,|\,v_{j}\in V_{T}\right\} },v_{k+1})\}).
\]
In other words, $v_{k+1}$ is added to $V_{T}$ and connected to its
earliest vertex, $v_{\min\left\{ i\,|\,v_{i}\in V_{T}\right\} }$,
which is defined as the \textbf{root} of $T$. It can be shown that,
if such a component exists, it is unique.
\item Otherwise, if there exists a connected component $T=(V_{T},E_{T})$
of $G_{k}$ containing a vertex $v=(t,x,y)$ such that:
\begin{itemize}
\item $x-d\leq x_{k+1}\leq x+d$ and $y-d\leq y_{k+1}\leq y+d$, and
\item $t_{k+1}-t\leq\delta$, 
\end{itemize}
then we define
\[
G_{k+1}:=(V_{k}\cup\{v_{k+1}\},E_{k}\cup\{(v_{\min\left\{ j\,|\,v_{j}\in V_{T}\right\} },v_{k+1})\}).
\]
If multiple connected components satisfy these conditions, one of
them is selected arbitrarily.
\item Otherwise, we simply add the vertex without introducing new edges:
\[
G_{k+1}:=(V_{k}\cup\{v_{k+1}\},E_{k}).
\]
\end{enumerate}
It follows that $G$ is a \textbf{polyforest}, i.e., a directed acyclic
graph (DAG) whose underlying undirected graph is a forest. Each connected
component $T=(V_{T},E_{T})$ forms a polytree and contains a distinguished
root vertex defined as

\[
v_{root}(T)=v_{\min\left\{ j\,|\,v_{j}\in V_{T}\right\} }\in V_{T}.
\]

The algorithm presented in this work aims to identify all such roots
corresponding to clusters $T$ satisfying $|V_{T}|\geq n$, for a
predefined threshold $n\geq3$. Equivalently, the primary output of
the algorithm is the subset of $V$ consisting of the roots of all
clusters meeting this size criterion. 

Event cameras often include noisy or overly sensitive pixels that
generate events at high rates due to internal noise, rather than genuine
changes in brightness. In practical applications, it is therefore
desirable to discard clusters arising from such spurious activity.
Since these pixels typically act in isolation, it is useful to impose
a constraint requiring that a valid cluster be supported by at least
a minimum number of distinct contributing pixels, denoted by a predefined
threshold $m$. As described below, the proposed algorithm naturally
incorporates this filtering mechanism within the event-processing
loop.

In light of the above, the complete output of the algorithm is represented
as a list, denoted $Clusters$, in which each entry $Clusters[j]$
contains the following fields:
\begin{itemize}
\item $Clusters[j][0]$ = the timestamp of the root (i.e., the initial timestamp
of the cluster),
\item $Clusters[j][1]$ = the x-coordinate of the root,
\item $Clusters[j][2]$ = the y-coordinate of the root,
\item $Clusters[j][3]$ = the timestamp of the last event related to the
cluster,
\item $Clusters[j][4]$ = the total number of events in the cluster,
\item $Clusters[j][5]$ = the number of distinct pixels contributing to
the cluster.
\end{itemize}
\textbf{Remark:} In this work, we present a version of the algorithm
that outputs only the roots of the clusters rather than the clusters
themselves. Nevertheless, for applications that require full cluster
reconstruction, it is straightforward to extend the method to build
the clusters during the event-processing loop without affecting the
$\Theta(N)$ time complexity. We adopt the present formulation for
two reasons. First, it highlights a novel perspective: extracting
key cluster properties without explicitly constructing the clusters.
Second, this formulation offers a clearer presentation of the core
idea underlying the algorithm.

\subsection{Demonstration of Algorithm Output }

\begin{figure}[H]
\includegraphics[width=4.7in]{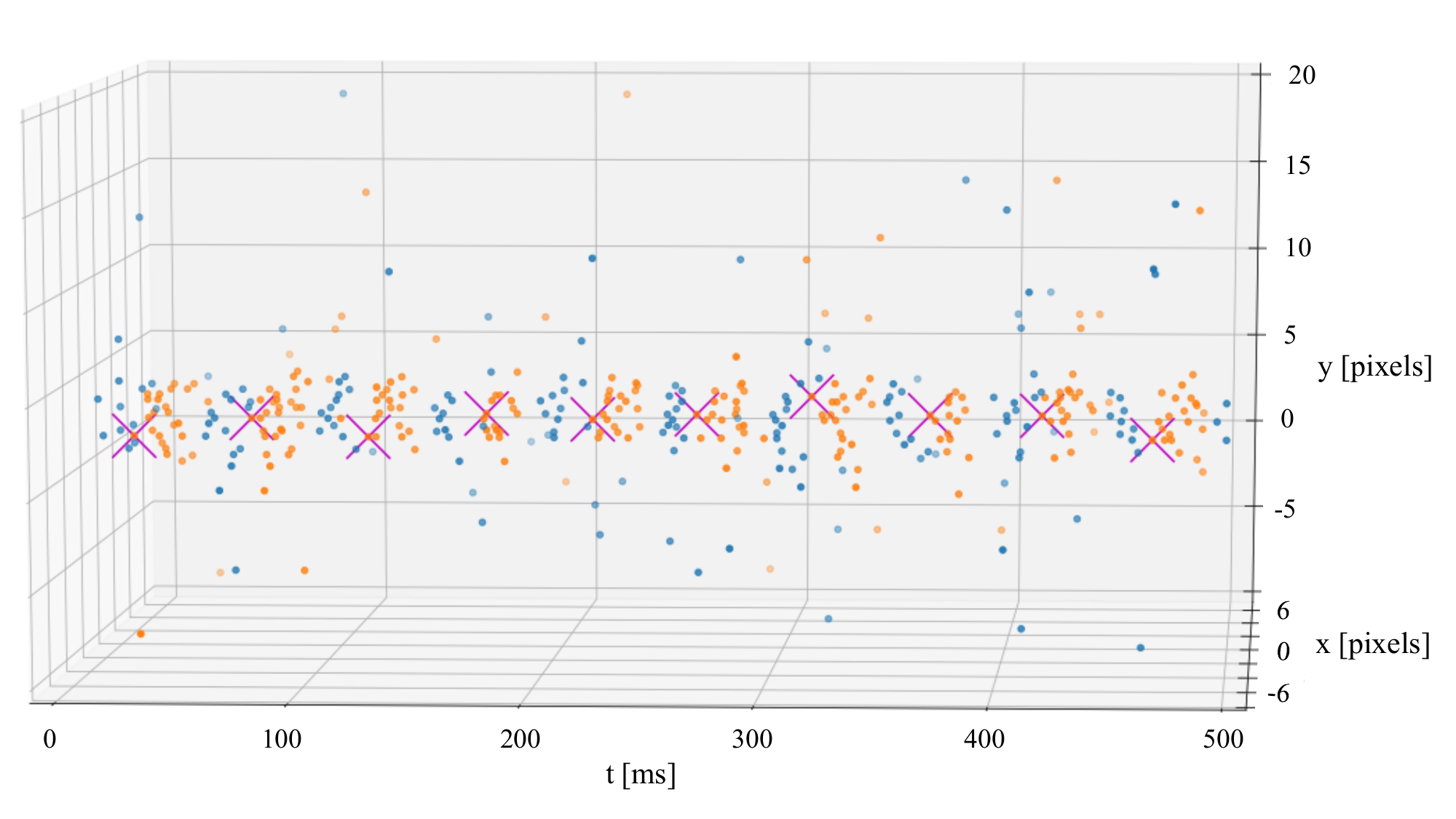}

\caption{Clusters corresponding to the periodic signal generated by the incandescent
lamp. The roots of the clusters formed by positive events are marked
with purple X symbols.\label{fig2-1}}
\end{figure}

In Figure \ref{fig2-1}, we present a 3D visualization of the roots
detected by the proposed clustering algorithm. For this demonstration,
we utilize data from a prior study by the authors \cite{key-7}, which
investigated a bias-tuning heuristic for optimizing the parameters
governing event generation in an event camera. The recording used
here corresponds to the configuration identified as optimal in that
work.

Recalling the experimental setup, we used an incandescent lamp powered
by a 50 Hz sinusoidal electrical grid, which effectively produces
a 100 Hz periodic signal of the form $\left|\sin(x)\right|$. The
signal was recorded using a Prophesee EVK4-HD event camera positioned
several meters from the lamp. To obtain a sparse signal, the lamp
was covered, leaving only a small aperture, resulting in only a few
dozen events per period.

Figure \ref{fig2-1} shows the events corresponding to ten periods
of the lamp signal. In the visualization, orange points represent
positive events (corresponding to increases in illumination), while
blue points represent negative events. For clarity, the figure displays
a 3D representation of the events within a small region of interest
(ROI) extracted from the camera\textquoteright s full megapixel array.

Using this dataset, the algorithm was applied to the positive events
with the following parameters: $\delta=2\,ms$, $d=1$, $n=10$, and
$m=5$. The algorithm successfully identified the root of each period
of the signal, corresponding to the first positive event within each
cluster. These roots are marked in the figure with purple X symbols.

\section{Algorithm Description\label{sec:Algorithm-description}}

\subsection{Constant Parameters}

As discussed in the introduction, the proposed algorithm is governed
by two primary criteria: the spatio-temporal distance between events
and the cluster size. Accordingly, it relies on four fixed parameters
that remain constant throughout its execution:
\begin{itemize}
\item $\delta$ = maximum temporal distance between adjacent events in a
cluster.
\item $d$ = maximum spatial distance (in pixels) between adjacent cluster's
events.
\item $n$ = minimum number of events per valid cluster. We assume $n\geq3$.
\item $m$ = minimum number of distinct pixels contributing to a valid cluster. 
\end{itemize}

\subsection{Dynamic arrays}

Let $r$ and $s$ denote the dimensions of the pixel array. The algorithm
maintains several $r\times s$ dynamic arrays, where each entry stores
information associated with the corresponding pixel, and is updated
during execution. 

The most basic array is $TimeSurface[u,v]$ which stores the timestamp
of the most recent event reported by pixel $(u,v)$, and is initialized
to $0$ . 

The remaining arrays are divided into two groups: those that store
information about the most recent cluster rooted at pixel $(u,v)$
(Table \ref{tab:1}), and those that store information about the most
recent cluster in which pixel $(u,v)$ participated (Table \ref{tab:2}).
In other words, Table \ref{tab:1} treats each pixel as a potential
cluster root, whereas Table \ref{tab:2} treats the pixel as associated
with a root that may differ from itself.

\begin{table}[h]
\begin{centering}
\begin{tabular}[b]{|l|>{\centering}m{1cm}|>{\raggedright}m{6.5cm}|}
\hline 
Array & Initial

value & Meaning\tabularnewline
\hline 
\hline 
$ClusterBegin[u,v]$ & 0 & Timestamp of the root of the most recent cluster rooted at pixel $(u,v)$.\tabularnewline
\hline 
$ClusterEnd[u,v]$ & 0 & Timestamp of the most recent event in the cluster rooted at $(u,v)$.\tabularnewline
\hline 
$grade[u,v]$ & 0 & Number of events in the most recent cluster rooted at $(u,v)$.\tabularnewline
\hline 
$pixels[u,v]$ & 0 & Number of distinct pixels contributing to the most recent cluster
rooted at $(u,v)$.\tabularnewline
\hline 
$ClusterId[u,v]$ & -1 & Index of the most recent cluster rooted at $(u,v)$.\tabularnewline
\hline 
\end{tabular}
\par\end{centering}
\caption{Dynamic arrays capturing the evolving properties of the most recent
cluster rooted at each pixel.\label{tab:1}}
\end{table}

\begin{table}[h]
\begin{centering}
\begin{tabular}[b]{|l|>{\centering}m{1cm}|>{\raggedright}m{6.5cm}|}
\hline 
Array & Initial

value & Meaning\tabularnewline
\hline 
\hline 
$PointerX[u,v]$ & -1 & Pointers to the root of the most recent cluster in which pixel $(u,v)$
participated.\tabularnewline
\cline{1-2} \cline{2-2} 
$PointerY[u,v]$ & -1 & (a value of $(-1,-1)$ indicates that the pixel is not associated
with any cluster).\tabularnewline
\hline 
$Compatibility[u,v]$ & 0 & Timestamp marking the beginning of the most recent cluster $(u,v)$
was related to.\tabularnewline
\hline 
\end{tabular}
\par\end{centering}
\caption{Dynamic arrays capturing the evolving properties of the clusters to
which each pixel is associated.\label{tab:2}}
\end{table}

\subsection{Algorithm block diagram}

\begin{figure}
\includegraphics[width=4.5in]{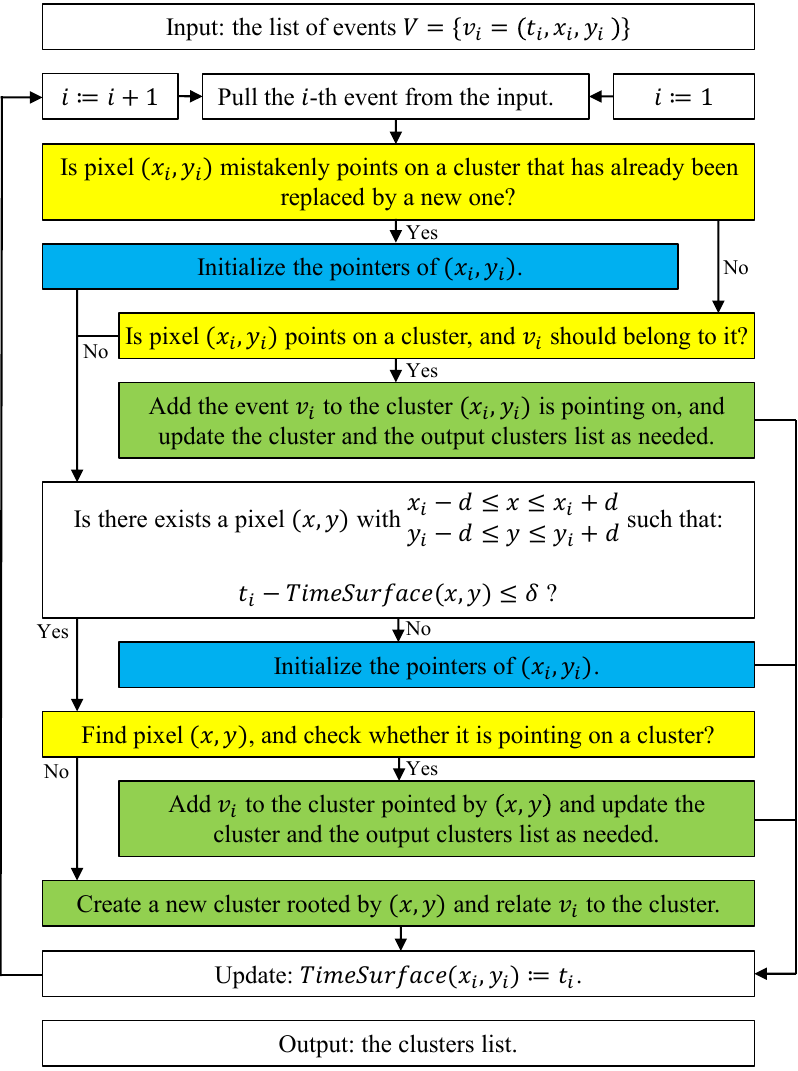} 

\caption{Algorithm block diagram.\label{fig1}}

\end{figure}

The diagram in Figure \ref{fig1} describes the decision-making process
of the algorithm. Several blocks in the diagram are color-coded to
represent different stages, which are explained in detail below. Yellow
blocks correspond to conditions evaluated by the algorithm, blue blocks
represent initialization steps, and green blocks denote operations
related to the generation and update of output clusters.

In the following, we consider an incoming event $v_{i}=(t_{i},x_{i},y_{i})$
from the input stream. Note that as mentioned above, the algorithm
does not make use of the event polarity. We now provide a detailed
explanation of the blocks in the diagram.

\subsubsection{Pointers initialization}

These operations are represented by the blue blocks. The initialization
of the pointers for pixel $(x_{i},y_{i})$ consists of the following
steps:
\begin{itemize}
\item $PointerX(x_{i},y_{i})=PointerY(x_{i},y_{i})=-1$.
\item $Compatibility(x_{i},y_{i})=0$.
\end{itemize}

\subsubsection{Conditions}

The block diagram contains three yellow blocks, each representing
a condition within the algorithm. In this subsection, we detail the
operations associated with these conditions.

\subparagraph*{First yellow block. }

The algorithm checks whether the pixel $(x_{i},y_{i})$ currently
points to a cluster root $(x,y)$, that has already been superseded
by a new cluster. This prevents mistakenly associating the event $v_{i}$
with an outdated cluster rooted at $(x,y)$. This requires the following
conditions:
\begin{itemize}
\item The pixel $(x_{i},y_{i})$ is considered to point to a cluster if
\begin{equation}
PointerX(x_{i},y_{i})\geq0.\label{eq: condition}
\end{equation}
Noting that both pointers are initialized to $-1$ when no cluster
is associated, one may use $PointerY(x_{i},y_{i})\geq0$, or require
both coordinates to be non-negative.
\item The cluster $(x_{i},y_{i})$ is pointing on is considered outdated
if 
\end{itemize}
\[
Compatibility(x_{i},y_{i})\neq ClusterBegin(PointerX(x_{i},y_{i}),PointerY(x_{i},y_{i})).
\]

\subparagraph*{Second yellow block. }

The algorithm checks whether pixel $(x_{i},y_{i})$ points to a cluster
(similar to the condition in Equation \ref{eq: condition}) and, if
so, whether 
\[
t_{i}-ClusterEnd(PointerX(x_{i},y_{i}),PointerY(x_{i},y_{i}))\leq\delta.
\]
If this condition holds, the event $v_{i}$ is associated with the
corresponding cluster.

\subparagraph*{Third yellow block. }

At this stage, it is guaranteed that there exists at least one $d$-neighborhood
pixel $(x,y)$ of $(x_{i},y_{i})$ such that 
\[
t_{i}-TimeSurface(x,y)\leq\delta.
\]
The algorithm then proceeds as follows:
\begin{itemize}
\item Select the pixel $(x,y)$ that minimizes $t_{i}-TimeSurface(x,y)$.
In the rare case of multiple candidates, ties are broken by selecting
the pixel corresponding to the most recent event, or by any fixed
rule.
\item Check whether the selected pixel $(x,y)$ is currently associated
with a cluster (see Equation \ref{eq: condition}). 
\end{itemize}

\subsubsection{Actions}

The diagram includes three green blocks, corresponding to the action
stages of the algorithm. The final block represents the creation of
a new cluster, while the other two correspond to cluster updates.

\subparagraph*{New cluster creation. }

In this case, there exists a pixel $(x,y)$ within a $d$-neighborhood
of $(x_{i},y_{i})$ such that 

\[
t_{i}-ClusterEnd(PointerX(x,y),PointerY(x,y))\leq\delta
\]
and $(x,y)$ is not currently associated with any cluster. The algorithm
then performs the following updates:
\begin{align*}
PointerX(x,y) & :=PointerX(x_{i},y_{i}):=x\\
PointerY(x,y) & :=PointerY(x_{i},y_{i}):=y\\
ClusterBegin(x,y) & :=TimeSurface(x,y)\\
Compatibility(x,y) & :=Compatibility(x_{i},y_{i}):=TimeSurface(x,y)\\
ClusterEnd(x,y) & :=t_{i}\\
grade(x,y) & :=2\\
pixels(x,y) & :=\begin{cases}
1 & (x,y)=(x_{i},y_{i})\\
2 & (x,y)\neq(x_{i},y_{i})
\end{cases}
\end{align*}

\subparagraph*{Cluster update. }

Updating a cluster requires modifying both the dynamic arrays and
the output list $Clusters$. Regarding the update of the dynamic arrays,
we distinguish between two cases:
\begin{enumerate}
\item \textbf{Extending an existing cluster via the same pixel.} In this
case, the event $v_{i}$ is connected to the cluster to which pixel
$(x_{i},y_{i})$ currently points. Let
\[
(P,Q):=(PointerX(x_{i},y_{i}),PointerY(x_{i},y_{i})).
\]
Then:
\begin{align*}
ClusterEnd(P,Q) & :=t_{i}\\
grade(P,Q) & :=grade(P,Q)+1
\end{align*}
\item \textbf{Attaching a pixel via a neighboring cluster.} In this case,
the new pixel $(x_{i},y_{i})$ is attached to the cluster referenced
by its $d$-neighborhood pixel $(x,y)$. Accordingly, the pointers
of $(x_{i},y_{i})$ must also be updated. Let
\[
(P,Q):=(PointerX(x,y),PointerY(x,y)),
\]
 and perform the following updates:
\begin{align*}
PointerX(x_{i},y_{i}) & :=P\\
PointerY(x_{i},y_{i}) & :=Q\\
Compatibility(x_{i},y_{i}) & :=Compatibility(x,y)\\
ClusterEnd(P,Q) & :=t_{i}\\
grade(P,Q) & :=grade(P,Q)+1\\
pixels(P,Q) & :=pixels(P,Q)+1
\end{align*}
\end{enumerate}

\subparagraph*{Updating the output list.}

After updating the cluster in the dynamic arrays, the output list
$Clusters$ is updated only if:
\[
grade(P,Q)\geq n\qquad\textrm{and}\qquad pixels(P,Q)\geq m.
\]

If these conditions are satisfied, two cases arise::
\begin{enumerate}
\item \textbf{Cluster already exists in the list.} This case corresponds
to the following two conditions:
\begin{align*}
ClusterId(P,Q) & >-1\\
t_{i}-Clusters[ClusterId(P,Q)][3] & \leq\delta
\end{align*}
In this case, we update:
\begin{align*}
Clusters[ClusterId(P,Q)][3] & :=ClusterEnd(P,Q)\\
Clusters[ClusterId(P,Q)][4] & :=grade(P,Q)\\
Clusters[ClusterId(P,Q)][5] & :=pixels(P,Q)
\end{align*}
\item \textbf{Cluster not yet in the list.} Otherwise, create a new entry:
\begin{align*}
ClusterId(P,Q) & :=\textrm{number\;of\;rows\;in\;Clusters}\\
Clusters[ClusterId(P,Q)][0] & :=ClusterBegin(P,Q)\\
Clusters[ClusterId(P,Q)][1] & :=P\\
Clusters[ClusterId(P,Q)][2] & :=Q\\
Clusters[ClusterId(P,Q)][3] & :=ClusterEnd(P,Q)\\
Clusters[ClusterId(P,Q)][4] & :=grade(P,Q)\\
Clusters[ClusterId(P,Q)][5] & :=pixels(P,Q)
\end{align*}
\end{enumerate}

\section{Conclusion}

This paper presented an asynchronous, event-driven clustering algorithm
for real-time detection of small event clusters in event camera data.
By leveraging the temporal structure of the event stream, the method
operates in a single pass without revisiting past events, achieving
linear time complexity $\Theta(N)$ independent of sensor resolution
and the number of clusters. A key feature of the proposed approach
is its ability to identify cluster roots online, immediately once
a predefined threshold is reached, making it well suited for low-latency
applications. Additionally, the algorithm incorporates inherent robustness
to noise through constraints on both event count and spatial support.
The graph-theoretic formulation provides a clear and principled interpretation
of the clustering process. Overall, the proposed method offers a simple
yet powerful framework for real-time event-based clustering, combining
theoretical clarity with practical efficiency, and is expected to
serve as a useful building block for a wide range of neuromorphic
vision applications.

\end{document}